\documentclass{article}

\usepackage[preprint]{neurips_2025}

\usepackage[utf8]{inputenc} 
\usepackage[T1]{fontenc}    
\usepackage{hyperref}       
\usepackage{url}            
\usepackage{booktabs}       
\usepackage{amsfonts}       
\usepackage{nicefrac}       
\usepackage{microtype}      
\usepackage{xcolor}         

\usepackage[inline]{enumitem}
\usepackage{mathtools}
\usepackage{amsthm}
\usepackage{amssymb}
\usepackage{dsfont}

\usepackage{adjustbox}
\usepackage{graphicx}
\usepackage{caption} 
\usepackage{subcaption}
\usepackage{wrapfig}
\usepackage{multirow}
\usepackage{makecell}
\usepackage{threeparttable}
\usepackage{multicol}

\theoremstyle{plain}
\newtheorem{theorem}{Theorem}[section]

\newtheorem{lemma}[theorem]{Lemma}
\newtheorem{corollary}[theorem]{Corollary}
\theoremstyle{definition}
\newtheorem{definition}[theorem]{Definition}

\usepackage{amsthm}

\usepackage[textsize=tiny]{todonotes}

\usepackage[linesnumbered, ruled, vlined]{algorithm2e}

\title{Integrated Influence: Data Attribution with Baseline}

\author{%
  Linxiao Yang, Xinyu Gu, Liang Sun\\
  DAMO Academy, Alibaba Group, Hangzhou, China\\
  \texttt{linxiao.ylx@alibaba-inc.com}, \texttt{guxinyue.gxy@alibaba-inc.com}, \\
  \texttt{liang.sun@alibaba-inc.com} \\
}

\begin{document}

\maketitle

\begin{abstract}
As an effective approach to quantify how training samples influence test sample, data attribution is crucial for understanding data and model and further enhance the transparency of machine learning models. We find that prevailing data attribution methods based on leave-one-out (LOO) strategy suffer from the local-based explanation, as these LOO-based methods only perturb a single training sample, and overlook the collective influence in the training set. On the other hand, the lack of baseline in many data attribution methods reduces the flexibility of the explanation, e.g., failing to provide counterfactual explanations. In this paper, we propose Integrated Influence, a novel data attribution method that incorporates a baseline approach. Our method defines a baseline dataset, follows a data degeneration process to transition the current dataset to the baseline, and accumulates the influence of each sample throughout this process. We provide a solid theoretical framework for our method, and further demonstrate that popular methods, such as influence functions,
can be viewed as special cases of our approach.
Experimental results show that Integrated Influence generates more reliable data attributions compared to existing methods in both data attribution task and mislablled example identification task. 
\end{abstract}


\section{Introduction}


Data attribution~\cite{koh2017understanding}, which aims to study the relationship between training data and model predictions, is essential for understanding the data and model, and building machine learning systems~\cite{hammoudeh2024training}. Data attribution enables practitioners to identify bias sources, enhance data quality, and improve model robustness, thus building more trustworthy and high-performing models. 
Furthermore, it plays a crucial role in achieving transparency and interpretability, which are crucial for building trust in machine learning systems across various applications. As the complexity and scale of machine learning models continue to grow, the ability to effectively attribute model performance to distinct training samples not only aids in refining models but also provides valuable guidance for future data collection and model development strategies.

These data attribution methods can be broadly categorized into retraining-based methods~\cite{feldman2020neural,kwon2022beta,wang2023data,ilyas2022datamodels} and gradient-based influence estimators~\cite{koh2017understanding,pruthi2020estimating,park2023trak}.
Among these gradient-based methods, the Influence Function (IF)~\cite{koh2017understanding} is a well-known influence estimator that analyzes how the model parameters change when the weight of a training sample $\boldsymbol{z}_i$ is infinitesimally perturbed by $\epsilon_i$. 
In fact, most data attributions are based on the Leave-One-Out (LOO)~\cite{ling1984residuals} strategy. Thus, these approaches are inherently local, concentrating on a very small neighborhood around the current training sample to estimate the sample importance and impact. Intuitively, these LOO-based methods only perturb a single sample, thus overlooking the collective influence in data distribution. 

\begin{figure*}[t]
\begin{center}
\includegraphics[width=0.32\columnwidth]{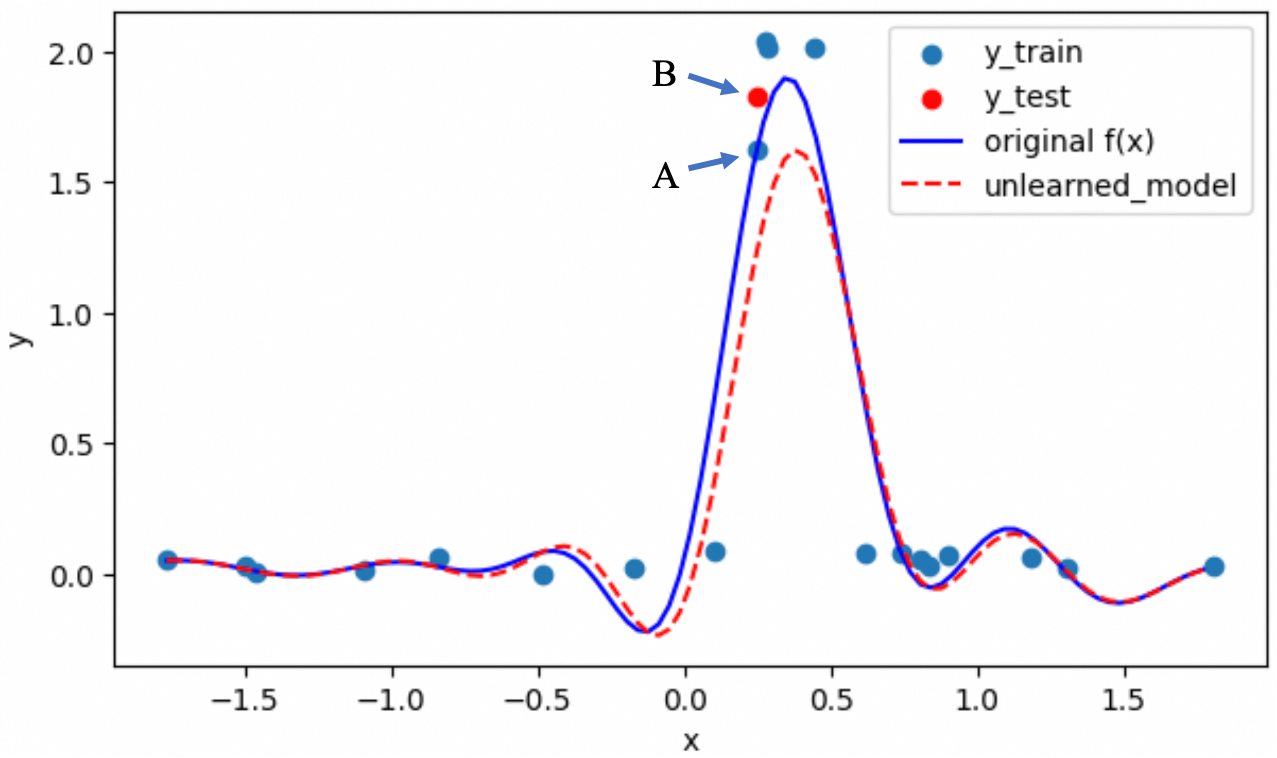}
\includegraphics[width=0.32\columnwidth]{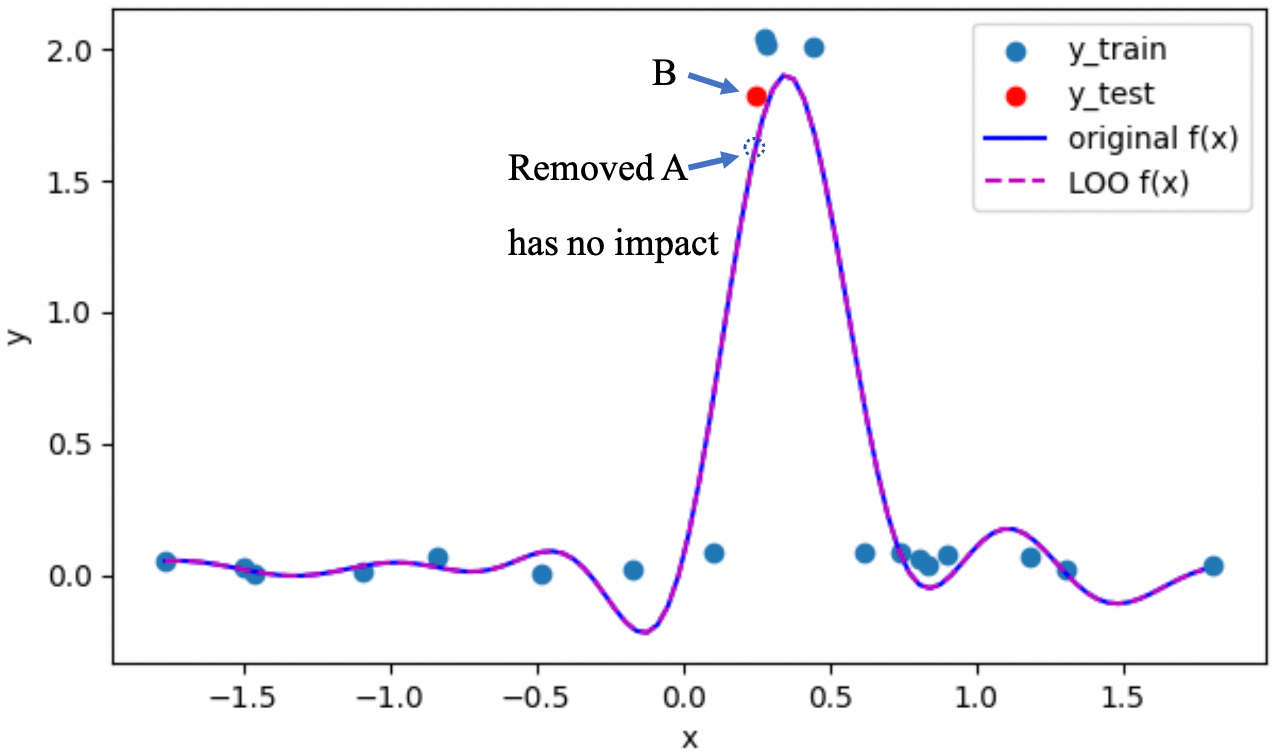}
\includegraphics[width=0.32\columnwidth]{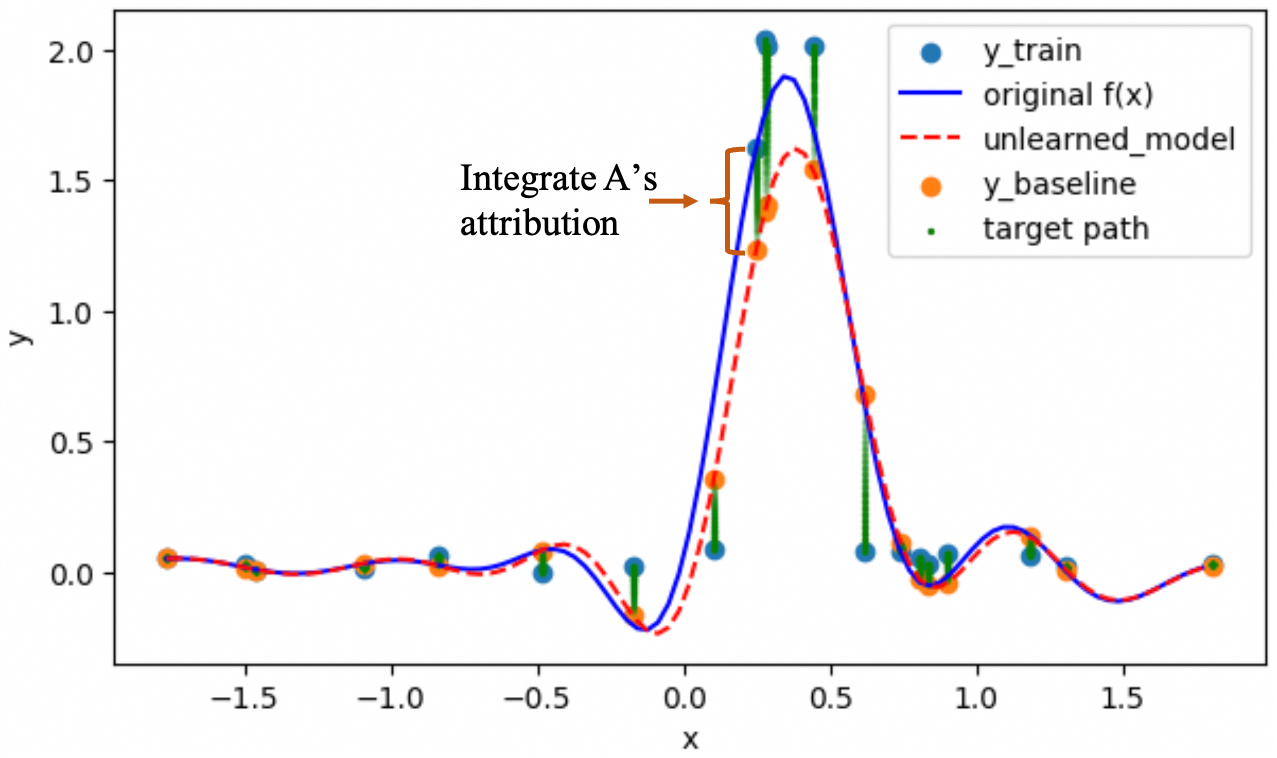}
\caption{Left: Toy example of a regression problem using the function $sinc$. Middle: Illustration of the counterintuitive phenomenon in the LOO approach, where the training sample A shows no contribution to the testing sample B, despite their proximity. Right: The baseline derived by our proposed method and the corresponding target path.}
\label{fig:if-bad-case}
\end{center}
\end{figure*}

To illustrate why LOO is considered a local method, let’s check a simplified example which involves a kernel regression problem. 
Fig.~\ref{fig:if-bad-case} illustrates a regression problem with the function $sinc$ as a kernel. The trained model is depicted by the blue line. We aim to evaluate how the training sample A, which the model fits perfectly, influences the predictive error of the test sample B, shown as a red circle in Fig.~\ref{fig:if-bad-case}. Following the LOO approach, we remove training sample A and retrain the model. The updated model, displayed as a red dashed line and labeled as LOO 
$f(x)$, remains unchanged from the original. Thus, it leads to the counterintuitive conclusion that removing sample A has no effect on the model and, consequently, no impact on sample B's predictive error, despite sample A being close to sample B.
This occurs because LOO's design inherently ignores contextual relationships: the triangular cluster of sample A with two adjacent points collectively supports predictions at B through spatial co-dependencies, but LOO's localized perturbation treats each sample as an independent unit. By construction, the method cannot detect how groups of nearby samples jointly shape predictions through their geometric configuration, thereby falsely attributing null influence to A despite its role in a shared predictive structure
Another limitation of the current data attribution methods is the lack of the baseline in explanation. 
In explainable AI (XAI), the baseline serves as a reference point, which also defines the context in which explanation is performed. The baseline can make the explanation more meaningful and enable comparisons. 
Many existing XAI methods such as SHAP~\cite{shap2017lee} and Integrated Gradients~\cite{integrated:gradient:2017} measures feature contribution relative to a baseline. For example, the baseline is the model's expected output in SHAP. 
In fact, baseline is also important in data attribution. Data attribution inherently seeks to answer “what if” counterfactual questions, which require comparing the current model to a reference baseline. 
By introducing the baseline in data attribution, we can answer more complex questions such as why a model predicts outcome A instead of outcome B. 
Unfortunately, most existing data attribution methods struggle to address such issues effectively.




We address these challenges through \textbf{Integrated Influence}, 
a novel data attribution framework that quantifies influence as the path integral from a non-informative baseline to the current training set. By gradually reintroducing information to the baseline dataset, we accumulate each point’s contribution across the trajectory, not just at the final model. Rigorously, we build a sold theoretical framework for our integrated influence. One byproduct of this framework is the explanation of why influence function is a local method. This approach naturally subsumes influence function as a special case: IF corresponds to infinitesimal perturbations near the current dataset. 
In summary, our contributions are:
\begin{itemize}
    \item  \textbf{Free of Locality Bias}: We build a novel data attribution framework based on a non-informative baseline, address the aforementioned challenges of existing data attribution methods. \vspace{-1mm}
    \item \textbf{Unification of Gradient Methods}: Our integrated influence is a general framework. We show influence function is a special case of our method, and TRAK can also be generalized from integrated influence by extending it to regression tasks. \vspace{-1mm}
    \item \textbf{Solid Theoretical Support}: We provide a solid theoretical foundation and framework for our integrated influence method. \vspace{-1mm}
    \item \textbf{Empirical Validation}: Our experiments in data attribution task and mislablled example identification task demonstrate the superior performance of our integrated influence to influence function, TracIn and TRAK.     
\end{itemize}


\section{Methodology}\label{sec:method}
\vspace{-2mm}
In this paper, we focus on supervised learning tasks, including both classification and regression.
Consider a training dataset $\mathcal{D}_{\text{train}}=\{\boldsymbol{z}_i\}_{i=1}^N$ comprising $N$ samples, where each sample $\boldsymbol{z}_i$ includes a feature vector $\boldsymbol{x}_i$ and a label vector $\boldsymbol{y}_i$.  Let $f(\cdot;\boldsymbol{\theta}^{*})$ be the model trained on $\mathcal{D}_{\text{train}}$ by minimizing the loss function $L^{\text{train}}(\mathcal{D}_{\text{train}})$, which is defined as the empirical expectation of the loss over each sample, i.e., $L^{\text{train}}(\mathcal{D}_{\text{train}}) = \mathbb{E}_{(\boldsymbol{x},\boldsymbol{y}) \in \mathcal{D}_{\text{train}}} l^{\text{train}}(f(\boldsymbol{x};\boldsymbol{\theta}), \boldsymbol{y})$. Here, $l^{\text{train}}(f(\boldsymbol{x};\boldsymbol{\theta}), \boldsymbol{y})$ measures the discrepancy between the prediction of the model $f(\boldsymbol{x};\boldsymbol{\theta})$ and the truth label at the data point $(\boldsymbol{x}, \boldsymbol{y})$, and $\boldsymbol{\theta}$ represents the model parameters.
Given a test sample $\boldsymbol{\hat{z}} = (\boldsymbol{\hat{x}}, \boldsymbol{\hat{y}})$, our aim is to attribute the performance of the model on $\boldsymbol{\hat{z}}$, i.e., $L^{\text{test}}(\boldsymbol{\hat{z}}; \boldsymbol{\theta}^{*})$, to the training samples in $\mathcal{D}_{\text{train}}$.

The contribution of a training sample \((\boldsymbol{x}_i, \boldsymbol{y}_i)\) arises from its unique information (exclusive patterns uncaptured by other samples) and shared information (latent correlations preserved across the dataset). To rigorously quantify this dual contribution, we evaluate the model’s predictive risk \(L^{\text{test}}(\boldsymbol{\hat{z}}; \boldsymbol{\theta}^{*})\) against a baseline \(L^{\text{test}}(\boldsymbol{\hat{z}}; \boldsymbol{\theta}')\) trained on a statistically sanitized dataset \(\overline{\mathcal{D}}_{\text{train}}\), designed to annihilate both information types. Crucially, conventional leave-one-out (LOO) methods—which remove \((\boldsymbol{x}_i, \boldsymbol{y}_i)\) from \(\mathcal{D}_{\text{train}}\)—only suppress unique information while inherently retaining shared information through co-dependent samples, thereby biasing attribution measurements. To resolve this, we propose constructing \(\overline{\mathcal{D}}_{\text{train}}\) not by sample removal, but through targeted perturbations that systematically erase unique information while decoupling shared dependencies, establishing a foundation for disentangled influence analysis.

\subsection{Integrated Influence}\label{subsec:integrated:inf}

By defining the baseline dataset, we can study the impact of each training sample by comparing the training set and the baseline dataset. Formally, given a sample $(\boldsymbol{x}, \boldsymbol{y})$, both the feature vector and the label can be modified to remove its unique information. To reduce the flexibility and simplify the discussion, we define the baseline dataset $\overline{\mathcal{D}}_{\text{train}}$ as having the same size and feature vectors as $\mathcal{D}_{\text{train}}$ but with different target values, i.e., $\overline{\mathcal{D}}_{\text{train}}=\{(\boldsymbol{x}_i,\overline{\boldsymbol{y}}_i)\}_{i=1}^N$. In other words, we restrict the baseline dataset identical to the training set in terms of the sample features, and the difference only lies in the target. We will later discuss the selection of $\overline{\boldsymbol{y}}_i$.

Given the training dataset $\mathcal{D}_{\text{train}}$ and the baseline dataset $\overline{\mathcal{D}}_{\text{train}}$, we analyze the impact of each training sample on the test samples by comparing the two datasets. Specifically, we consider data samples move from  the baseline dataset back to the original training dataset. This process can be understood as gradually reintroducing unique information to $\overline{\mathcal{D}}_{\text{train}}$ to observe how it affects the performance on the test samples.
To facilitate this analysis, we introduce the concept of a Dataset Path, which defines the process of transiting data from $\overline{\mathcal{D}}_{\text{train}}$ to $\mathcal{D}_{\text{train}}$. Since the only differences between $\mathcal{D}_{\text{train}}$ and $\overline{\mathcal{D}}_{\text{train}}$ are in the target values, we first define the notion of a Target Path, which outlines how to gradually adjust the target value of a baseline sample $\overline{\boldsymbol{z}}_i$ to match that of the corresponding training sample $\boldsymbol{z}_i$.

\begin{definition}[Target Path]\label{def:target_path}
Let $\mathcal{M}$ represent a continuous manifold on which all target vectors of the training dataset $\mathcal{D}_{\text{train}}$ and the baseline dataset $\overline{\mathcal{D}}_{\text{train}}$ reside. Consider the target vectors for the $i$-th sample in these datasets, denoted as $\boldsymbol{y}_i$ from $\mathcal{D}_{\text{train}}$ and $\overline{\boldsymbol{y}}_i$ from $\overline{\mathcal{D}}_{\text{train}}$. A target path, $\rho_i$, is defined as a continuous and smooth function that maps a scalar $t$ to a vector on the manifold $\mathcal{M}$, satisfying the boundary conditions $\rho_i(0) = \overline{\boldsymbol{y}}_i$ and $\rho_i(1) = \boldsymbol{y}_i$.
\end{definition}

Note that in the target path we only consider the transition for the target. The definition of target path establishes a smooth transition for the target vector from its baseline value to its original value within the dataset using the continuous manifold $\mathcal{M}$.
In the following, we extend the concept of a path between two data points to that of two datasets.

\begin{definition}[Dataset Path]
Given a training dataset $\mathcal{D}_{\text{train}}=\{(\boldsymbol{x}_i,\boldsymbol{y}_i)\}_{i=1}^N$
and a baseline dataset $\overline{\mathcal{D}}_{\text{train}}=\{\boldsymbol{x}_i,\overline{\boldsymbol{y}}_i\}_{i=1}^N$.
A dataset path $\Gamma(t)$ is defined as a function
that maps scalar $t$ to a dataset $\{(\boldsymbol{x}_i,\rho_i(t))\}_{i=1}^N$,
where $\rho_i$ is target path between $\overline{\boldsymbol{y}}_i$ and $\boldsymbol{y}_i$. Clearly, we have
$\Gamma(0)=\overline{\mathcal{D}}_{\text{train}}$, $\Gamma(1)=\mathcal{D}_{\text{train}}$.
\end{definition}

Intuitively, the dataset path connects two datasets, providing infinite intermediate datasets that allow us to analyze how the transition from the baseline dataset to the training dataset affects the test samples. It is evident that such a path is not unique. We will discuss how to select the dataset path later.

Based on the dataset path defined above, we can measure the effect of model change due to the change of training data on the test sample $\boldsymbol{\hat{z}} = (\boldsymbol{\hat{x}}, \boldsymbol{\hat{y}})$. Specifically, we can analyze how the gradual transition from the baseline dataset $\overline{\mathcal{D}}_{\text{train}}$ to the training dataset $\mathcal{D}_{\text{train}}$ affects the model's performance on the test sample along the data path. Formally, given the training dataset $\mathcal{D}_{\text{train}}$, baseline dataset $\overline{\mathcal{D}}_{\text{train}}$,
and dataset path $\Gamma(t)$ which transits smoothly between the two datasets, the difference of loss function on the test sample $\boldsymbol{\hat{z}}$ (denoted as $\Delta$) can be formulated as
\begin{align}
\Delta=L^{\text{test}}(\boldsymbol{\hat{z}},\boldsymbol{\theta}^*) - L^{\text{test}}(\boldsymbol{\hat{z}},\boldsymbol{\theta}')
= \int_{0}^{1}\frac{\partial L^{\text{test}}(\boldsymbol{\hat{z}},\boldsymbol{\theta}(\Gamma(t)))}{\partial t}dt,
\end{align}
where $\boldsymbol{\theta}(\Gamma(t))$ denotes the optimal parameters learned from the dataset $\Gamma(t)$, i.e.,
\begin{align}
    \boldsymbol{\theta}(\Gamma(t))=\arg\min_{\boldsymbol{\theta}} \mathbb{E}_{(\boldsymbol{x},\rho(t)) \in \Gamma(t)} l^{\text{train}}(f(\boldsymbol{x};\boldsymbol{\theta}), \rho(t)). 
\end{align}
This integral formulation allows us to understand how incremental changes along the dataset path affect the test performance. Here, $\frac{\partial L^{\text{test}}(\boldsymbol{\hat{z}},\boldsymbol{\theta}(\Gamma(t)))}{\partial t}$ represents the rate of change of the performance measure $L^{\text{test}}$ as we move gradually from the baseline dataset to the training dataset.

Intuitively, this decomposition provides insights into the sensitivity of the model to different aspects of the data and how each incremental change contributes to the overall difference in performance.
We reformulate the integral as follows:
\begin{align}
\Delta
=\int_{0}^{1}\frac{\partial L^{\text{test}}(\boldsymbol{\hat{z}},\boldsymbol{\theta}(\Gamma(t))}{\partial \boldsymbol{\theta}(\Gamma(t))}\frac{\partial\boldsymbol{\theta}(\Gamma(t))}{\partial t}dt. \label{eqn:delta}
\end{align}
To derive the expression for $\frac{\partial\boldsymbol{\theta}(\Gamma(t))}{\partial t}$, we recall that $\boldsymbol{\theta}(\Gamma(t))$ is optimal for each
$t$, giving us:
\begin{align} 
\frac{\partial\boldsymbol{\theta}(\Gamma(t))}{\partial t}=&-\boldsymbol{H}(t)^{-1}\frac{\partial ^2L^{\text{train}}(\Gamma(t))}{\partial\boldsymbol{\theta}\partial t},
\label{eqn:ptheta_pt}
\end{align}
where $\boldsymbol{H}(t)$ denotes the Hessian matrix
of $L^{\text{train}}$ with respect to $\boldsymbol{\theta}$.
Given the decomposition structure of $L^{\text{train}}$, and
only the labels differ among the datasets $\Gamma(t)$, then we have
\begin{align} 
\frac{\partial ^2L^{\text{train}}(\Gamma(t))}{\partial\boldsymbol{\theta}\partial t}=\sum_{i}\frac{\partial ^2l^{\text{train}}(f(\boldsymbol{x};\boldsymbol{\theta}), \rho_i(t))}{\partial\boldsymbol{\theta}\partial t}
=\sum_{i=1}^N\frac{\partial ^2l^{\text{train}}(f(\boldsymbol{x};\boldsymbol{\theta}), \rho_i(t))}{\partial\boldsymbol{\theta}\partial \rho_i(t)}\frac{\partial\rho_i(t)}{\partial t}.
\label{eqn:jacc}
\end{align}
Substituting Eq.\eqref{eqn:ptheta_pt} and Eq.~\eqref{eqn:jacc} into Eq.~\eqref{eqn:delta}, then we arrive at
\begin{align} 
\Delta=-\sum_{i=1}^N\int_{0}^{1}\boldsymbol{G}(t)\boldsymbol{H}^{-1}(t)\boldsymbol{J}_i(t)\frac{\partial\rho_i(t)}{\partial t}dt, \label{eq:contribution:int}
\end{align}
where 
\begin{align}
    \boldsymbol{G}(t)=&\frac{\partial L^{\text{test}}(\boldsymbol{\hat{z}},\boldsymbol{\theta}(\Gamma(t))}{\partial \boldsymbol{\theta}(\Gamma(t))},\\
    \boldsymbol{J}_i(t)=&\frac{\partial ^2l^{\text{train}}(f(\boldsymbol{x};\boldsymbol{\theta}), \rho_i(t))}{\partial\boldsymbol{\theta}\partial \rho_i(t)}. 
\end{align}
In the above equation, the term $\boldsymbol{H}^{-1}\boldsymbol{J}_i(t)$
represents how a slight modification of the target of the $i$-th sample affects the optimal parameters of the model.
Meanwhile, $\boldsymbol{G}(t)$ describes how these optimal parameters impact the performance on the test sample $\boldsymbol{\hat{z}}$.
Thus, the entire integrand reflects how changes in the $i$-th sample's target influence the performance measurement on the testing sample. 

Motivated by Eq.~\eqref{eq:contribution:int}, we can quantify the impact of the $i$-th sample on the testing performance on $\boldsymbol{\hat{z}}$ by considering how even slight changes in its target values adjust the model's parameters and consequently influence the test results. By integrating these effects over the path from the baseline to the training dataset, we gain a comprehensive view of the sample's contribution to performance. Thus, we term our proposed method as Integrated Influence, which is formally defined below. 

\begin{definition}[Integrated Influence]
We define the contribution of the $i$-th sample introduced by integrated influence as:
\begin{align}
I(i)=-\int_{0}^{1}\boldsymbol{G}(t)\boldsymbol{H}^{-1}(t)\boldsymbol{J}_i(t)\frac{\partial\rho_i(t)}{\partial t}dt. \label{eqn:imp-define}
\end{align} 
And the total contribution on the test sample $\boldsymbol{\hat{z}}$ is $\Delta=\sum_{i=1}^N I(i)$. 
\end{definition}

Direct computation of the integral in $I(i)$ is intractable, as it involves finding the optimal model parameters for datasets along the dataset path $\Gamma(t)$. To reduce computational complexity, we sample $K$ datasets from the path and approximate the integral using the Euler method. Specifically, given the sampled datasets $\{\Gamma(t_k)\}_{k=0}^K$, with $\Gamma(t_K)=\mathcal{D}_{\text{train}}$ and $\Gamma(t_0)=\overline{\mathcal{D}}_{\text{train}}$, we approximate $I(i)$ as follows:

\begin{definition}[Discrete Integrated Influence]
We define the discrete approximate contribution of the $i$-th sample introduced by integrated influence as:
\begin{align}
    I(i)\approx -\sum_{k=1}^{K}\boldsymbol{G}(t_k)\boldsymbol{H}^{-1}(t_k)\boldsymbol{J}_i(t_k)\boldsymbol{\Delta}_y^{(k)},\label{def:iif}
\end{align}
where
$\boldsymbol{\Delta}_y^{(k)}=(\rho_i(t_k)-\rho_i(t_{k-1}))$.    
\end{definition}

By discretizing the integral, we can efficiently estimate $I(i)$ by accumulating the influence of each training sample on the sampled datasets $\{\Gamma(t_k)\}_{k=1}^K$. Our experimental results demonstrate that with just a few datasets, i.e., a small $K$, we can achieve an accurate estimation of $I(i)$.


\subsection{Relation to Influence Function}\label{subsec:relation}
In this subsection, we discuss the relationship between our integrated influence and the widely used Influence Function~\cite{koh2017understanding}, highlighting their differences, particularly when considering scenarios where $L^{\text{train}}$  is the mean square error (MSE) loss.

First, we examine the relationship between our method and the Influence Function in general cases, as formalized in the following lemma:

\begin{lemma}\label{lemma:if-general}
The Influence Function can be viewed as a special case of Eq.~\eqref{eqn:imp-define} by assuming that $\boldsymbol{G}(t)$ and $\boldsymbol{H}(t)$ are independent of $t$, where the baseline dataset is set as $\overline{\mathcal{D}}_{\text{train}}=\{(\boldsymbol{x}_i,\boldsymbol{y}_i')\}_{i=1}^{N}$, with $\boldsymbol{y}_i'$ being a target value that satisfies $\frac{\partial l^{\text{train}}(f(\boldsymbol{x};\boldsymbol{\theta}), \boldsymbol{y}')}{\partial\boldsymbol{\theta}}=0$.
Moreover, in scenarios where $\frac{\partial l^{\text{train}}(f(\boldsymbol{x};\boldsymbol{\theta}), \boldsymbol{y})}{\partial\boldsymbol{\theta}}$ is linear with respect to $\boldsymbol{y}$, such as when $l^{\text{train}}$ is MSE loss or Cross-entropy loss, the Influence Function is exactly equals Eq.~\eqref{def:iif} with $K=1$, and baseline dataset is set to $\overline{\mathcal{D}}_{\text{train}}=\{(\boldsymbol{x}_i,\boldsymbol{y}_i')\}_{i=1}^{N}$.
\end{lemma}

Although the Influence Function does not explicitly introduce a baseline dataset, Lemma~\ref{lemma:if-general} indicates that it implicitly defines a baseline, rendering it a special case of our method. The rationale behind the baseline described in Lemma~\ref{lemma:if-general} is that the Influence Function removes the effect of the $i$-th sample by adjusting its target values such that the sample contributes nothing to the overall gradient of the dataset with respect to the parameters.

To further understand the baseline introduced by the Influence Function, we consider a simplified scenario where $L^{\text{train}}$ is the MSE loss, and the target $\boldsymbol{y}$ reduces to a scalar, denoted as $y$. Given the loss function $l^{\text{train}}(f(\boldsymbol{x}_i,\boldsymbol{\theta}), \rho
_i(t)) = (f(\boldsymbol{x},\boldsymbol{\theta}) - \rho_i(t))^2$, the influence function of the $i$-th sample is:

\begin{align}
\frac{\partial L^{\text{test}}(\boldsymbol{\hat{z}}, \boldsymbol{\theta}^*)}{\partial \boldsymbol{\theta}^*} \boldsymbol{H}^{-1} \frac{\partial f(\boldsymbol{x}, \boldsymbol{\theta}^*)}{\partial \boldsymbol{\theta}^*} (y_i - f(\boldsymbol{x}_i, \boldsymbol{\theta}^*)).\label{eqn:if-mse}
\end{align}



By comparing the influence function with our method, we arrive at the following conclusion:
\begin{corollary}\label{lemma:if-mse}
For MSE loss, the Influence Function can be viewed as a special case of Eq.~\eqref{def:iif} with $K=1$, where the baseline dataset is defined as $\overline{\mathcal{D}}_{\text{train}} = \{(\boldsymbol{x}_i, f(\boldsymbol{x}_i, \boldsymbol{\theta}^*))\}_{i=1}^{N}$.
\end{corollary}

This analysis reveals that under MSE loss, the Influence Function implicitly assumes a baseline dataset wherein the target values are replaced by the model's predictions, thereby nullifying their influence on the parametrization.
Nevertheless, this analysis also reveals a counterintuitive phenomenon: if the $i$-th sample has no fitting error, meaning its model prediction equals its true target value, the Influence Function concludes that this sample has no contribution to any testing samples. 
This analysis explains the phenomenon presented in Fig.~\ref{fig:if-bad-case}.
We note that this phenomenon is not confined to the MSE loss but exists for any loss function where $\frac{\partial l^{\text{train}}(f(\boldsymbol{x}_i,\boldsymbol{\theta}), f(\boldsymbol{x}_i,\boldsymbol{\theta}))}{\partial \boldsymbol{\theta}} = 0$. In these scenarios, the Influence Function fails to assign the correct importance score to points with no fitting errors. This observation highlights the critical role of baseline selection: an improper baseline can lead to suboptimal attribution results. To address this problem, we introduce a method for selecting an appropriate baseline using the concept of unlearning in Section~\ref{subsec:unlearning}. 

\vspace{-0.2cm}
\subsection{Finding Baseline Dataset using Unlearning}\label{subsec:unlearning}
Our objective is to construct a baseline dataset that simultaneously disrupts both unique and shared information components within training samples. A naive baseline approach involves either assigning all targets in the original training set \(\mathcal{D}_{\text{train}}\) to a predefined constant or performing random label permutations across samples.
However, such a dataset would inherently be non-informative, significantly altering the original model, leading to a longer path, and increasing approximation error.


Our key insight is that each sample typically contains two types of information: one that influences the test sample's error and another that is orthogonal, or irrelevant, to the test sample. Therefore, when analyzing the $i$-th sample in the training set, we need to carefully modify its target value to ensure that it does not retain information useful for the testing sample. This effectively renders the sample excluded from influencing the test outcome.
Building on this insight, we propose a baseline generation method based on machine unlearning~\cite{guo2020certified,nguyen2022survey}. 
Machine unlearning~\cite{machine:unlearning:2021} is the process of removing the influence of specific data points from a trained machine learning model without retraining from scratch. In this paper, for the given test sample \(\boldsymbol{\hat{z}}\), we perform unlearning by maximizing its predictive loss, which is equivalent to minimizing $-L^{\text{test}}(\boldsymbol{\hat{z}}; \boldsymbol{\theta})$. 
To prevent catastrophic forgetting and keep the baseline target close to the original target, we incorporate a regularization term that minimizes the training loss, ensuring the model retains information from the training samples. The model parameters are then updated to 
$\boldsymbol{\theta}'$ by solving the following optimization problem:
\begin{align}
\boldsymbol{\theta}'=\arg\min_{\boldsymbol{\theta}}
    -L^{\text{test}}(\boldsymbol{\hat{z}}; \boldsymbol{\theta})+\lambda L^{\text{train}}(\mathcal{D}^{\text{train}}),\label{eqn:unlearning}
\end{align}
where $\lambda>0$ is a parameter controlling the regularization. 
With the updated parameters $\boldsymbol{\theta}'$, the model effectively unlearns information about the test data, allowing us to use its predictions to construct a baseline dataset. The appropriate baseline dataset is then defined as:
\begin{align}
    \overline{\mathcal{D}}_{\text{train}}=\{(\boldsymbol{x}_i,f(\boldsymbol{x}_i,\boldsymbol{\theta}'))\}_{i=1}^N.\label{eqn:baseline}
\end{align}

Note that solving the optimization problem in Eq.~\eqref{eqn:unlearning} can be time-consuming and impractical in real-world scenarios. To mitigate this challenge, we observe that it is sufficient to update the network for only a few epochs without the need for full convergence to the optimal solution.

\vspace{-0.2cm}
\subsection{Dataset Path Generation}
After finding the baseline dataset, next we discuss how to generate a dataset path that bridges the original training dataset and the baseline dataset.
When the target values in the baseline dataset closely resemble those in the original dataset, the manifold \(\mathcal{M}\) encompassing the target vectors of both the training dataset \(\mathcal{D}_{\text{train}}\) and the baseline dataset \(\overline{\mathcal{D}}_{\text{train}}\) can be approximately considered a linear space. In this neighborhood of the manifold, we can use a linear function to approximate. Thus, we choose a linear path as follows:
\begin{align}
    \rho_i(t) = t\boldsymbol{y}_i+(1-t)f(\boldsymbol{x}_i,\boldsymbol{\theta}').
\end{align}
In its discrete form, this is expressed as:
\begin{align}
    \rho_i(t_k)=\frac{k}{K}\boldsymbol{y}_i+\left(1-\frac{k}{K}\right)f(\boldsymbol{x}_i,\boldsymbol{\theta}').\label{eqn:path}
\end{align}
With dataset path defined above, we can train a model for each
sampled dataset. Given $K$ sampled dataset, we need 
train $K-1$ models. To reduce the computational complexity, we initialize the model parameters with the current ones and assume that the model can be optimized to its optimal state within one epoch of Stochastic Gradient Descent (SGD). Formally, we set $\boldsymbol{\theta}(\Gamma(t_K))=\boldsymbol{\theta}^*$, and for $1\le k<K$, we have 
\begin{align}
    \boldsymbol{\theta}(\Gamma(t_k)) = \boldsymbol{\theta}(\Gamma(t_{k+1})) - \eta\frac{\partial L^{\text{train}}(\Gamma(t_k))}{\partial \boldsymbol{\theta}},
    \label{eqn:path-model}
\end{align}
where $\eta$ is the learning rate.
By updating the model $K-1$ times, we efficiently obtain $K-1$ models.

\vspace{-0.2cm}
\subsection{Accelerating via Hessian Matrix Compression}
The computation of Hessian matrix usually is the computational burden. One trick to accelerate the computation is to project the Hessian matrix as well as the gradient matrix to a low-dimensional subspace, which has been used in TRAK~\cite{park2023trak} and is proven to be efficient. Specifically,
let $\boldsymbol{A}\in \mathbb{R}^{M\times P}(P\ll M)$ be a matrix with full column rank.
Then $\boldsymbol{A}\boldsymbol{A}^{\dagger}$ forms a $P$-dimension subspace, where $\boldsymbol{A}^{\dagger}$ denotes the pseudo-inverse of matrix $\boldsymbol{A}$.
Then we have following approximation 
\begin{align}
\boldsymbol{H}^{-1}(t_k)
\approx &\boldsymbol{A}\boldsymbol{A}^{\dagger}\boldsymbol{H}^{-1}(t_k)\boldsymbol{A}^{\dagger}\boldsymbol{A}
=\boldsymbol{A}(\boldsymbol{A}^T\boldsymbol{H}(t_k)\boldsymbol{A})^{-1}\boldsymbol{A}.\label{eqn:compression}
\end{align}
The rational behind such approximation is that for some large model, Hessian matrix can be approximated by a low-rank approximation, thus we can compress it before inverse it. With this approximation, we only need to inverse a matrix of size $P\times P$ with the cost reducing to $\mathcal{O}(P^3)$.
The computational complexity can be further reduced
by directly computing the inverse Hessian-vector product
instead of computing the matrix inversion. 
This further reduces the 
computational complexity to $\mathcal{O}(P^2)$.

\vspace{-3mm}
\section{Experiments}
\vspace{-2mm}
We show the performance of our proposed data attribution method in data attribution tasks and mislabelled example identification tasks. The hardware configuration is described in detail in the Appendix.

\textbf{Baselines.}
We compare our method with three state-of-the-art methods, namely Influence Function (IF)~\cite{koh2017understanding}, TracIn~\cite{pruthi2020estimating}, and TRAK~\cite{park2023trak}. Both Influence Function and TRAK are derived to approximate the LOO estimator. Similar to our method, TRAK compresses the gradients using random projection, making it suitable for relatively large models. In contrast, the Influence Function requires computing the Inverse-Hessian-Vector-Product, which demands substantial memory when dealing with models with a large number of parameters.
To address the memory requirements, we implement the Influence Function in real-world experiments using its variant with Conjugate Gradients~\cite{dengdattri} to accelerate the method. To make the fair and consistent comparison, we use the implementation from \cite{dengdattri}  for all three baselines. We denote our proposed Integrated Influence method as IIF.

\textbf{Evaluation Metric.} In the task of data attribution, attribution methods assign importance scores to each training sample. The quality of these importance scores is evaluated by measuring their ability to make counterfactual predictions. Specifically, we assess these methods using the Linear Datamodeling Score (LDS), which is defined as the Spearman rank correlation between the true loss of the model with a subset of training samples and the estimated loss using the importance scores, i.e., \(\text{corr}(\boldsymbol{p}, \boldsymbol{q})\). Here, \(\text{corr}(\cdot, \cdot)\) denotes the Spearman rank correlation operation.
The vectors \(\boldsymbol{p}\) and \(\boldsymbol{q}\) are defined as follows: the \(i\)-th elements of \(\boldsymbol{p}\) and \(\boldsymbol{q}\) are given by \(L^{\text{test}}(\boldsymbol{z}, \boldsymbol{\theta}_{S_i})\) and \(\sum_{j \in S_i} I(j)\), respectively. The set \(S_i\) is a randomly generated index set, and \(\boldsymbol{\theta}_{S_i}\) denotes the model parameters learned using the training samples whose indices are in \(S_i\).
By correlating these vectors, we can effectively evaluate the capability of the attribution method to assign meaningful importance scores that align with actual model performance changes.
To compute the LDS for the data attribution task, we randomly select 5000 independent index sets. For each of these index sets, a model is trained, resulting in a total of 5000 models being trained. This extensive modeling allows for a robust evaluation of how well the importance scores can predict the impact of different subsets of training data on the model's performance.

For the task of identifying mislabeled examples, we use the Area Under the Receiver Operating Characteristic curve (AUC) to measure the performance of each method in identifying these mislabeled instances. 

\vspace{-3mm}
\subsection{Experiments on Data Attribution}\label{sec:data-attri}
\vspace{-2mm}
We begin by conducting experiments in a linear regression task using synthetic data. We first generate feature vectors $\boldsymbol{x}_i$ for 100 training samples. Each sample contains 10 features, with feature values drawn from an independent and identically distributed (iid) normal distribution.
Next, a 10-dimensional weight vector $\boldsymbol{w}$ is independently drawn from a normal distribution. The target value is computed as $y_i=\boldsymbol{x}_i^T\boldsymbol{w}+\epsilon_i^n$, where $\epsilon_i^n$ represents white noise with a mean of zero and a standard deviation of $\sigma_n$. The test samples are generated using the same procedure as the training samples, but the variance of the noise added to the target is set to $\sigma_s$.

We perform linear regression with $L^{\text{train}}$ set to the MSE loss and conduct data attribution using respective methods. We also set $L^{\text{test}}$ to the MSE loss. Notably, in this setting, both the baseline presented in Eq.~\eqref{eqn:baseline} and the path shown in Eq.~\eqref{eqn:path} have closed-form solutions. This setting also perfectly matches the assumptions of the influence function, where the loss function with respect to the parameters is convex. As TARK is designed for classification problems, it is not included in these experiments.

\begin{figure*}[t]
\begin{center}
\includegraphics[width=0.95\columnwidth]{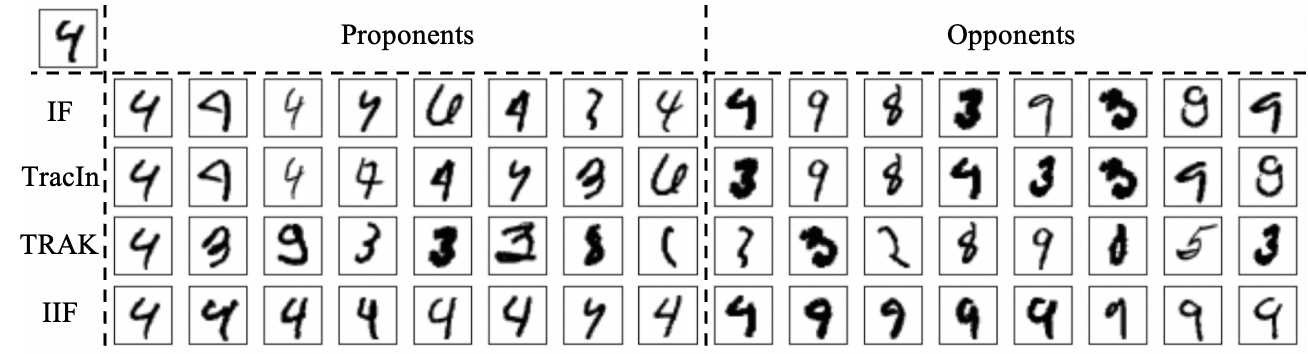}
\vspace{-3mm}
\caption{Top 8 proponents and opponents examples obtained by different methods on the MNIST dataset. The test sample is shown in top-left corner.}
\label{fig:pro_opp}
\end{center}
\vspace{-0.5cm}
\end{figure*}

\begin{table}[t]
\centering
\caption{LDS of different methods under different noise variance settings.}
\vspace{-3mm}
    \label{tab:lr-noise-distribution}
\begin{tabular}{ c | c | c | c }
\hline 
& \makecell{$\sigma_n=0.1$\\$\sigma_s=1$} & \makecell{$\sigma_n=1$\\$\sigma_s=1$} & \makecell{$\sigma_n=1$\\$\sigma_s=0.1$} \\
\hline
\hline
IIF & \textbf{0.859}& \textbf{0.744}& \textbf{0.768}\\
\hline
IF & 0.856& 0.634& 0.423\\
\hline
TracIn & 0.815& 0.607& 0.405\\
\hline
\end{tabular}\label{table:LR:LDS}
\vspace{-6mm}
\end{table}

We conduct experiments to evaluate the performance of various data attribution methods under different noise levels. Specifically, we assess the quality of the data importance scores generated by each method using LDS. Note that when the training data has low noise (i.e., $\sigma_n/\sigma_s$ is small), models trained with randomly selected subsets of the training samples tend to have similar performances, resulting in comparable testing errors. In contrast, when the training data is relatively noisy (i.e., $\sigma_n/\sigma_s$ is large), the performance of models trained on different portions of the data can vary significantly. Thus, if a data attribution method inaccurately estimates the contribution of the data, the LDS will decrease substantially.
We vary the noise levels in the training dataset and testing dataset, respectively, considering three scenarios: $\sigma_n=0.1$ and $\sigma_s=1$,  $\sigma_n=1$ and $\sigma_s=1$, and $\sigma_n=1$ and $\sigma_s=0.1$. The LDS for each method is reported in Table~\ref{table:LR:LDS},
with results averaged over 100 independent trials.
From the table, we observe that when the noise level in the training data is low (i.e., $\sigma_n=0.1$), our proposed method and the Influence function exhibit similar LDS, slightly outperforming TracIn. However, as the noise level increases (i.e., $\sigma_n=1$ and $\sigma_n=1$), all three methods experience a decline in performance. Despite that, the proposed method shows relatively robust performance, outperforming the Influence function and TracIn by an increased margin as noise increases.
Additionally, the table indicates that the performance of the Influence function and TracIn is also affected by the noise level in the testing dataset. Their LDS values decrease even when the testing dataset noise is low (i.e., $\sigma_n=0.1$). In contrast, as the noise level in the testing dataset decreases, the LDS for our proposed method slightly increases. This experiment demonstrates that the proposed method outperforms the Influence function and TracIn, especially when the noise level in the training data is high.

We conduct experiments on a classification problem using a real-world dataset, specifically the MNIST dataset~\cite{lecun1998gradient}, which contains 10 classes. We randomly select 5000 samples to form the training dataset and 1000 samples for the testing dataset. We train a 4-layer Multi-Layer Perceptron (MLP) using cross-entropy loss. For consistency, we also use cross-entropy loss to measure the model's performance on the testing dataset.
In our method, we transform the original target labels into one-hot encoding to ensure the target values reside on a continuous manifold. We select baselines according to Eqs~\eqref{eqn:unlearning}, and \eqref{eqn:baseline}, minimizing the loss over 5 epochs using the SGD optimizer. We note that as our method accumulate the data importance along the path.

\begin{table}[t]
\centering
\caption{LDS of different methods on the MNIST dataset.}
\vspace{-3mm}
    \label{tab:lds-mnist}
\begin{tabular}{ c | c | c | c}
\hline 
 IF & TracIn & TRAK  & IIF \\
\hline
\hline
 0.1035 & 0.1379 & 0.1115  &\textbf{0.1635}\\
\hline
\end{tabular}
\vspace{-4mm}
\end{table}

We compute the data importance using each data attribution, as summarized in Table~\ref{tab:lds-mnist}. It shows that estimations from all methods are positively correlate with the loss of the true retrained model. Furthermore, our method achieves the highest LDS, outperforming the second-best method by approximately 18\%.

\vspace{-3mm}
\subsection{Visualization of Proponents and Opponents}
\vspace{-2mm}
In this experiment, we visualize the proponents and opponents of a test sample in the MNIST dataset to demonstrate our integrated influence method. 
We randomly select a test sample, which is an image of a ``4'' but  resembles a ``9''. This image is shown in the top-left corner of Fig.~\ref{fig:pro_opp}. Unlike other methods, we employ different baselines to identify proponents and opponents. For proponents, we update the model parameters to reduce the probability of the image belonging to the class ``4''. We use the predictions of this adjusted model on the training samples as baseline targets. By integrating the influence along the path, we calculate the importance of each sample using our proposed methods, during the integration process, the loss of the test sample is gradually decreasing. Samples with the highest negative importance, meaning they contribute most to the reduction of loss, are identified as proponents. Conversely, for opponents, we first decrease the loss, i.e. updating the model to increase the probability of the image being classified as a ``4''. Again, the model's predictions on the training samples serve as baseline targets. We integrate the influence along the path (the loss on testing samples increase gradually) and compute the importance of each sample using our proposed methods.
Samples with the highest positive importance, meaning they contribute most to preventing the decrease of the loss, are identified as opponents.
We present the top-8 proponents and opponents examples obtained by respective methods in Fig.~\ref{fig:pro_opp}. From the figure, it is evident that our method consistently identifies proponents visually similar to the test example. In contrast, our method is capable of selecting pixel-wise similar but differently labeled samples as opponents (all images in the opponents set are ``9'' and resemble a ``4'').


\vspace{-3mm}
\subsection{Identification of Mislabelled Examples}\label{sec:mislabel-identification}
\vspace{-2mm}
We conduct experiments in the task of identifying mislabeled examples using two real-world datasets: MNIST and CIFAR-10. For each dataset, we randomly select 1000 samples as the training datasets and randomly flip 10\% of the labels. We train a 4-layer MLP for the MNIST dataset and a ResNet-9 on the CIFAR-10 dataset. Our goal is to identify mislabeled samples using various data attribution methods.

\begin{table}[t]
\centering
\caption{AUC of different methods on MNIST and CIFAR-10 datasets. ``OOM" means out of memory for IF on the CIFAR-10 dataset.}
\vspace{-3mm}
    \label{tab:auc}
\begin{tabular}{c | c | c | c | c }
\hline 
& IF & TracIn & TRAK & IIF \\
\hline
\hline
MNIST &0.869 & 0.853 & 0.866 & \textbf{0.885}\\
\hline
CIFAR-10 &\text{OOM} & 0.677 & 0.728 & \textbf{0.736}\\
\hline
\end{tabular}
\vspace{-6mm}
\end{table}

Intuitively, the relationship between features and labels in mislabeled examples differs from that in correctly labeled examples. Moreover, since the mislabeled examples are randomly selected, their feature-label relationships vary among themselves as well. Consequently, each mislabeled sample's contribution to its training error, referred to as self-influence, is most significant. We use the negative self-influence of each sample as an estimate of the unnormalized probability that the sample is mislabeled. To compute self-influence, the testing samples are identical to the training samples.

In this case, identifying baseline datasets using Eqs~\eqref{eqn:unlearning} and \eqref{eqn:baseline} is not feasible. Instead, we determine a baseline for each individual sample. For the $i$-th training sample $(\boldsymbol{x}_i,\boldsymbol{y}_i)$, we find its baseline as $(\boldsymbol{x}_i, f(\boldsymbol{x}_i,\boldsymbol{\theta}_i')$, where
$
\boldsymbol{\theta}_i'=\boldsymbol{\theta}^*+\eta\frac{\partial l^{\text{train}}(f(\boldsymbol{x}_i,\boldsymbol{\theta}),\boldsymbol{y}_i)}{\partial \boldsymbol{\theta}}$.

Performance is measured using the AUC. The results for each method are presented in Table~\ref{tab:auc}. From the table, we observe that all methods achieve similar AUCs, with our method achieving the highest AUC compared to the others.

\vspace{-.3cm}
\section{Conclusion}
\vspace{-.3cm}

This paper presents integrated influence, a novel data attribution method that overcomes the limitations of LOO strategy by incorporating a baseline dataset. This approach enhances explanation flexibility, allowing for counterfactual analysis. 
We demonstrate the theoretical soundness of our method and show that it generalizes popular methods such as influence functions. Experimental results indicate that Integrated Influence provides more reliable data attributions, outperforming existing methods in tasks including data attribution and mislabeled example identification.
In the future, we plan to explore integrated influence in more real-world applications, and also develop and open source a more user-friendly toolkit. 

\newpage
\bibliographystyle{plain}
\bibliography{example_paper}

\newpage
\appendix
\section{Algorithm~\ref{algorithm-overall}}
We summarize the full algorithm in Algorithm~\ref{algorithm-overall}.

\RestyleAlgo{ruled}
\SetKwInOut{Input}{Input}\SetKwInOut{Output}{Output}
\SetKw{Define}{Define}
\SetKw{Return}{Return}
\SetKw{Initialize}{Initialize}
\begin{algorithm}[t]
\fontsize{9}{10}\selectfont
\caption{Integrated Influence (IIF)}
\label{algorithm-overall}
\Input{Training samples $\{(\boldsymbol{x}_i,\boldsymbol{y}_i)\}_{i=1}^N$, testing sample $\boldsymbol{\hat{z}}$, number of accumulated steps $K$,
		regularization parameter $\lambda$, learning rate $\eta$, compression matrix $\boldsymbol{A}$.}
\Output{Contribution of training samples $\{I(i)\}_{i=1}^N$.}
Unlearn the information of testing samples from the model using Eq.~\eqref{eqn:unlearning}, and generate the baseline dataset according to Eq.~\eqref{eqn:baseline}\;
Initialize $I(i) = 0,\quad \forall i=1,\dots,N$\;
\For{$k = K,...,1$}{
Generate sampled dataset $\Gamma(t_k)$ using Eq.~\eqref{eqn:path}\;
Compute $\boldsymbol{\theta}(\Gamma(t_k))$ using $\Gamma(t_k)$ according to Eq.\eqref{eqn:path-model}.\;
\For{$i = 1,...,N$}{
Compute $\boldsymbol{u}_i(t_k)$
and $\boldsymbol{J}_i(t_k)$\;
}
Compute the gradient matrix $\boldsymbol{G}(t_k)$\;
\For{$i = 1,...,N$}{
Compute sample contribution $I^{(k)}(i)$, according to Eqs.\eqref{def:iif} and \eqref{eqn:compression}\;
$I(i) = I(i)+I^{(k)}(i)$\;
}
}
\Return $I(i),\quad \forall i=1,\dots,N$\;
\end{algorithm}

\section{Proofs.}
\subsection{Proof of Lemma~\ref{lemma:if-general}}

\begin{proof}
First, we demonstrate the general case. By assuming $\boldsymbol{G}(t)$ and $\boldsymbol{H}(t)$ are constant, we have:
\begin{align}
    I(i)=&-\int_{0}^{1}\boldsymbol{G}(t)\boldsymbol{H}^{-1}(t)\boldsymbol{J}_i(t)\frac{\partial\rho_i(t)}{\partial t}dt\\
    =&-\boldsymbol{G}\boldsymbol{H}^{-1}\int_{0}^{1}\boldsymbol{J}_i(t)\frac{\partial\rho_i(t)}{\partial t}dt.
\end{align}

It can be further simplified as
\begin{align}
        I(i)=-\boldsymbol{G}\boldsymbol{H}^{-1}\left(\frac{\partial l^{\text{train}}(f(\boldsymbol{x};\boldsymbol{\theta}), \rho_i(t))}{\partial\boldsymbol{\theta}}\bigg|_{t=0}^1\right).
\end{align}

Given that $\rho_i(1)=\boldsymbol{y}_i$ and $\rho_i(0)=\boldsymbol{y}_i'$, this leads to

\begin{align}
I(i) = -\boldsymbol{G}\boldsymbol{H}^{-1} \left(\frac{\partial l^{\text{train}}(f(\boldsymbol{x};\boldsymbol{\theta}), \boldsymbol{y}_i)}{\partial \boldsymbol{\theta}} - \frac{\partial l^{\text{train}}(f(\boldsymbol{x};\boldsymbol{\theta}), \boldsymbol{y}_i')}{\partial \boldsymbol{\theta}}\right).
\end{align}

Since the baseline definition states that the latter derivative is zero, we obtain:

\begin{align}
I(i) = -\boldsymbol{G}\boldsymbol{H}^{-1} \frac{\partial l^{\text{train}}(f(\boldsymbol{x}; \boldsymbol{\theta}), \boldsymbol{y}_i)}{\partial \boldsymbol{\theta}}.
\end{align}

This is equivalent to the influence function, thus proving the general case.

Now, consider scenarios where $\frac{\partial l^{\text{train}}(f(\boldsymbol{x};\boldsymbol{\theta}), \boldsymbol{y}_i)}{\partial\boldsymbol{\theta}}$ is linear in $\boldsymbol{y}_i$. This implies:

\begin{align}
\boldsymbol{J}_i(t_k)\boldsymbol{\Delta}_y^{(k)} &= \frac{\partial l^{\text{train}}(f(\boldsymbol{x}; \boldsymbol{\theta}), \rho_i(t_k))}{\partial \boldsymbol{\theta}} \\
&- \frac{\partial l^{\text{train}}(f(\boldsymbol{x}; \boldsymbol{\theta}), \rho_i(t_{k-1}))}{\partial \boldsymbol{\theta}}. 
\end{align}

With $K=1$, we have:

\begin{align}
\boldsymbol{J}_i(t_K)\boldsymbol{\Delta}_y^{(K)} = \frac{\partial l^{\text{train}}(f(\boldsymbol{x}; \boldsymbol{\theta}), \boldsymbol{y}_i)}{\partial \boldsymbol{\theta}}.
\end{align}

Thus, referring to Eq.~\eqref{def:iif}, we have:

\begin{align}
    I(i) &\approx -\boldsymbol{G}(t_K)\boldsymbol{H}^{-1}(t_K)\boldsymbol{J}_i(t_K) \boldsymbol{\Delta}_y^{(K)} \\
    &= -\boldsymbol{G}\boldsymbol{H}^{-1} \frac{\partial l^{\text{train}}(f(\boldsymbol{x}; \boldsymbol{\theta}), \boldsymbol{y}_i)}{\partial \boldsymbol{\theta}}.
\end{align}

This conforms with the formulation of the Influence Function, completing our proof.
\end{proof}

\subsection{Proof of Lemma~\ref{lemma:if-mse}}
With the MSE loss, the Jacobian matrix of $l^{\text{train}}(f(\boldsymbol{x}_i,\boldsymbol{\theta}), \rho
_i(t))$ with respect to $\rho_i(t)$ and $\boldsymbol{\theta}$ is expressed as:
\begin{align}
    \frac{\partial^2 l^{\text{train}}(f(\boldsymbol{x}_i,\boldsymbol{\theta}), \rho
_i(t))}{\partial \boldsymbol{\theta} \partial \rho
_i(t)} = -\frac{\partial f(\boldsymbol{x}_i,\boldsymbol{\theta})}{\partial \boldsymbol{\theta}}.
\end{align}
This leads to the contribution of the $i$-th sample being given by:
\begin{align}
I^{(k)}(i) \approx \sum_{k=1}^K\boldsymbol{G}(t_k) \boldsymbol{H}^{-1}(t_k)  \frac{\partial f(\boldsymbol{x},\boldsymbol{\theta})}{\partial \boldsymbol{\theta}} \boldsymbol{\Delta}_y^{(k)}.
\end{align}
By setting $K=1$, we have
\begin{align}
I^{(k)}(i) \approx \sum_{k=1}^K\boldsymbol{G}\boldsymbol{H}^{-1}\frac{\partial f(\boldsymbol{x},\boldsymbol{\theta})}{\partial \boldsymbol{\theta}}(\boldsymbol{y}_i-\boldsymbol{y}'_i),  
\end{align}
where $\boldsymbol{y}'_i$ denotes the target vector in baseline dataset.
By comparing the above equation with (\ref{eqn:if-mse}),
we find that by setting $\boldsymbol{y}'_i=f(\boldsymbol{x}_i,\boldsymbol{\theta}^*)$, our method becomes an estimator same to the Influence function.

\begin{figure*}[t]
\begin{center}
\includegraphics[width=0.95\columnwidth]{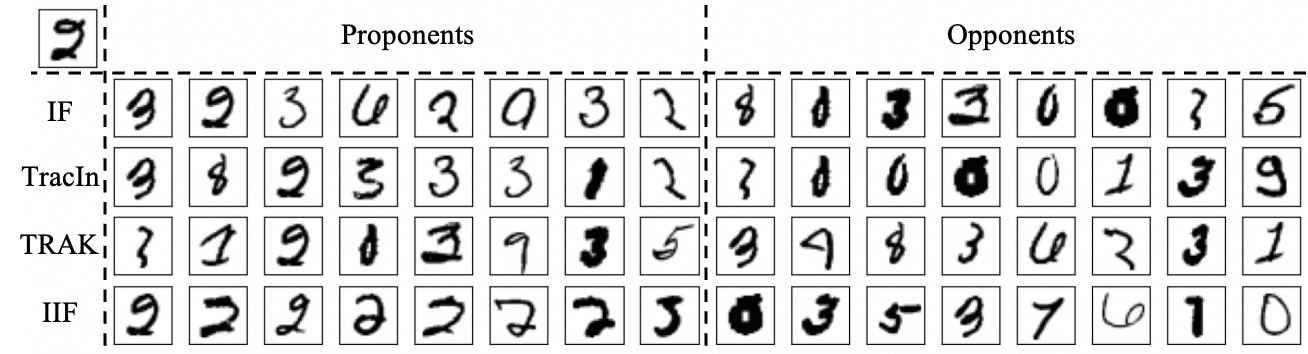}
\caption{Top 8 proponents and opponents examples obtained by respective methods.
The testing sample is shown in top-left corner.}
\end{center}
\end{figure*}
\section{Additional experimental results}
We also evaluate the performance of each method under different noise distributions. Specifically, we test three scenarios: (a) the noise in the training dataset follows a Gaussian distribution while the noise in the testing dataset follows a Laplacian distribution, denoted as "N\&L"; (b) the noise in the training dataset follows a Laplacian distribution while the noise in the testing dataset follows a Gaussian distribution, denoted as "L\&N"; and (c) the noise in both the training and testing datasets follows a Laplacian distribution, denoted as "L\&L."
When the noise follows a Laplacian distribution, using linear regression with a least squares estimator may present a model mismatch. This scenario tests the robustness of the methods to such mismatches. The results are reported in Table~\ref{tab:lr-noise-distribution}.
From the table, we observe that our method demonstrates robustness to model mismatch in both the training and testing samples.
Nevertheless, influence function and TracIn exhibit sensitivity to model mismatch in the training or testing data. 
Specifically, their LDS values decrease when model mismatch involves,
especially when the noise in the training data is set to Laplacian. 
Additionally, Table~\ref{tab:lr-noise-distribution} reveals that each method performs better when the noise in both the training and testing datasets follows a Laplacian distribution, compared to the scenario where the noise in the training dataset follows a Laplacian distribution and the noise in the testing dataset follows a Gaussian distribution.

\begin{table}[t]
\centering
\caption{LDS of respective methods under different noise distribution settings.}
    \label{tab:lr-noise-distribution}
\begin{tabular}{ c | c | c | c }
\hline 
& N\&L & L\&N & L\&L \\
\hline
\hline
IIF & \textbf{0.789}& \textbf{0.767}& \textbf{0.801}\\
\hline
IF & 0.688& 0.571& 0.628\\
\hline
TracIn & 0.654& 0.550& 0.601\\
\hline
\end{tabular}
\end{table}

\section{Additional Experimental Setting}
We conduct all experiments on a computer equipped with a 12-core CPU and a V100 GPU with 32GB of memory.

\textbf{Configure of 4-layer MLP and ResNet-9.} The configurations of 4-layer MLP and ResNet-9
are summarized in 
Table~\ref{tab:config-mlp}. 

\begin{table}[h]
\centering
\caption{Configuration of 4-layer MLP and ResNet-9.}
\scriptsize
\begin{tabular}{ c | c | c}\label{tab:config-mlp}
Parameters & MLP & ResNet-9\\
\hline
\hline
Latent size & [128,64,10]&[64,128,128,128,256,512,512,512,10]\\
\hline
Drop out &0.1  &0.1\\
\hline
Optimizer 
&\makecell{SGD\\lr=0.01,\\momentum=0.9} 
&\makecell{Adam\\lr=0.001} 
\\
\hline
epoch &50 & 10 \\
\end{tabular}
\end{table}

\textbf{Parameter Settings.} Table~\ref{tab:config-lambda} summarizes the settings for the regularization parameter $\lambda$ (as described in Equation (\ref{eqn:unlearning})) and the number of columns in the compression matrix (as described in Equation (\ref{eqn:compression})) for each task. It is important to note that no compression is applied in the case of linear regression task.

\begin{table}[t]
\centering
\caption{The setting of $\lambda$ and $P$ in each task.}
    \label{tab:config-lambda}
\small
\begin{tabular}{ c | c | c | c | c }
\hline 
& \multicolumn{2}{c|}{Data attribution} & \multicolumn{2}{c}{\makecell{Identification of \\Mislabelled Examples}}\\
\cline{2-5}
& Linear regerssion & MNIST & MNIST & CIFAR-10\\
\hline
\hline
$\lambda$ & 1 & 0.5 & 0 & 0\\
$\lambda$ & - & 256 & 256 & 256\\
\hline
\end{tabular}
\end{table}

\section{Implement Details}
\textbf{Compressing the Hessian Matrix.} 

\textbf{Replcing the Hessian Matrix with  Fisher Information Matrix.} 
In an optimal model, the Hessian matrix of the loss function with respect to model parameters should be non-negative definite. However, computational precision issues can lead to ill-conditioning in the Hessian, complicating its inversion. Notably, for optimal models, the expected Hessian equals the Fisher Information Matrix, when the loss $l^{\text{train}}$ is the negative log-likelihood. We can thus approximate the Hessian with the Fisher Information Matrix, known for being non-negative definite:
\begin{align}
    \boldsymbol{H}(t_k)\approx\sum_{i=1}^{N}\boldsymbol{u}_i(t_k)\boldsymbol{u}_i(t_k)^T,
    \label{eqn:hessian_approxi}
\end{align}
where $\boldsymbol{u}_i(t_k) = \frac{\partial l^{\text{train}}(f(\boldsymbol{x}_i,\boldsymbol{\theta}), \rho
_i(t_k))}{\partial\boldsymbol{\theta}}$.
This approach helps avoid issues related to computational precision and ensures the matrix negative definiteness. 

\textbf{Dealing with Classification Problems.}
For classification tasks with discrete labels, target vectors 
$\boldsymbol{y}$ are typically represented as probability distributions over classes using one-hot encoding. However, the baseline in Eq.~\eqref{eqn:baseline} results in dense vectors along the target path, which can incur computational challenges.
To illustrate, consider a binary classification problem with cross-entropy loss. Suppose we have an example with a true label $[0, 1]$ and a model prediction $[1-p, p]$. The cross-entropy loss for this scenario is $-\log p$, and its gradient with respect to the network parameters $\boldsymbol{\theta}$ is $-\frac{1}{p} \frac{\partial p}{\partial \boldsymbol{\theta}}$.
If the true label is modified to a dense vector, such as $[0.1, 0.9]$, the cross-entropy loss becomes $-0.9\log p - 0.1\log(1-p)$, and the gradient becomes $-\left(\frac{0.9}{p} - \frac{0.1}{1-p}\right) \frac{\partial p}{\partial \boldsymbol{\theta}}$. For a well-fitted sample where $p \rightarrow 1$, this gradient becomes a large number, leading to computational challenges.
To address this issue, we enforce sparsity in the target vectors along the target path when dealing with classification problems by applying $\rho_i(t) = \rho_i(t) \cdot \boldsymbol{y}_i$. This trick helps prevent large gradient magnitudes and maintains numerical stability.
\section{Related Work}

Training data influence analysis and data attribution~\cite{hammoudeh2024training},study the relationship between training data and model predictions by determining how to apportion credit for specific model behavior (especially on the test dataset) to the training instances. It can be roughly categorized into retraining-based methods and gradient-based influence estimators. In this paper, we focus on the latter class, including the influence function~\cite{koh2017understanding},TracIn~\cite{pruthi2020estimating}, TARK~\cite{park2023trak},  RelatedIF~\cite{barshan2020relatif}, GR~\cite{tsai2024sample}, SOIF~\cite{basu2020second}, DataModel~\cite{ilyas2022datamodels}, etc.

The influence function~\cite{koh2017understanding} is one of the best-known influence estimators. It considers how the model changes if the weight of the training instance/sample $z_i$ is infinitesimally perturbed by $\epsilon_i$. The influence function needs to calculate the inverse Hessian matrix directly, which is computationally prohibitive. To speed up the computation, \cite{koh2017understanding} proposes to estimate Hessian-vector product (HVP). The obvious advantage of the influence function is that it does not require to retrain the model. In particular, it assumes that the Hessian matrix is positive definite, which may not be hold for deep models.

TracIn~\cite{pruthi2020estimating} extends the influence function by training the training set fixed, and considers the change in model parameters as a function to time (or training iterations). Instead of relying on the Hessian matrix, it uses gradients computed throughout the training process. The key assumption is that training points with similar gradients to a test point will have a higher influence on its prediction. Although it avoids Hessian computations, TracIn sacrifices precision in influence estimation compared to traditional influence functions. Its key assumption is that gradient similarity is a good proxy for influence, which may not always hold.

\end{document}